# TRAFFIC SCENE RECOGNITION BASED ON DEEP CNN AND VLAD SPATIAL PYRAMIDS

**FANG-YU WU[1], SHI-YANG YAN[1], JEREMY S. SMITH[2], BAI-LING ZHANG[1]**

[1]Department of Computer Science and Software Engineering, Xi'an Jiaotong-Liverpool University, Suzhou, 215123, China
[2]Department of Electrical Engineering and Electronics, University of Liverpool, Liverpool, L69 3BX, UK
E-MAIL: fangyu.wu13@student.xjtlu.edu.cn

**Abstract:**

Traffic scene recognition is an important and challenging issue in Intelligent Transportation Systems (ITS). Recently, Convolutional Neural Network (CNN) models have achieved great success in many applications, including scene classification. The remarkable representational learning capability of CNN remains to be further explored for solving real-world problems. Vector of Locally Aggregated Descriptors (VLAD) encoding has also proved to be a powerful method in catching global contextual information. In this paper, we attempted to solve the traffic scene recognition problem by combining the features representational capabilities of CNN with the VLAD encoding scheme. More specifically, the CNN features of image patches generated by a region proposal algorithm are encoded by applying VLAD, which subsequently represent an image in a compact representation. To catch the spatial information, spatial pyramids are exploited to encode CNN features. We experimented with a dataset of 10 categories of traffic scenes, with satisfactory categorization performances.

## 1. Introduction

Humans have the remarkable ability to categorize complex traffic scenes very accurately and rapidly, which is important for the inference of the traffic situation and subsequent navigation in the complex and varying driving environment. It will be major achivement to implement an automatic traffic scene recognition system which imitates the human capability to understand traffic scenes. Such a system will play a crucial role toward the success of numerous applications, such as self-driving car/driverless car, traffic mapping and traffic surveillance [1]. Automatic acquisition of information from real-world traffic scenes will also be pivotal to optimize current traffic management system, for example, by improving traffic flow during busy periods [1].

Image representation has been studied for more than two decades, with a number of efficient hand-designed algorithms previously proposed for feature extraction. Among them, bag-of-features (BOF) methods represent an image as bags of locally extracted visual features, such as HoG (Histogram of Oriented Gradient) [2] and SIFT (Scale Invariant Feature Transform) [3]. Despite some limited success, hand-crafted features cannot reflect the rich variabilities hidden in the data. In recent years, Convolutional Neural Networks (CNN) [3][4] have brought breakthroughs in learning image representations. By training multiple layers of convolutional filters in an end-to-end network, CNNs are capable of detecting complex features automatically, which is a prerequisite for many of the computer vision tasks such as scene recognition. In many benchmark examples like image classification with the ImageNet dataset, superior performance have been reported comparing to earlier work which relied on hand-crafted features [5].

CNN features learnt from training data may contain much redundant information, from which a more compact representation could be achieved by using some feature coding schemes, for example, VLAD (Vector of Locally Aggregated Descriptors) [7] and Fisher Vectors (FV) [8], which have demonstrated tremendous successes in image processing tasks, e.g., image retrieval. Among these encoding strategies, VLAD has gained more popularity, with many excellent application examples such as scene recognition and object detection [9]. However, a well-known problem of VLAD is the absence of spatial layout information. In the BOF framework, the predominant approach to compensate for this drawback is to include spatial information by a Spatial Pyramid (SP) [4]. Following the same line of thought，we firstly build spatial pyramids of the images which are matched to region level CNN features, and then perform VLAD encoding on the separate pyramids. The generated VLAD codes are concatenated into a final representation, which is subsequently forwarded to a classifier, e.g. a SVM, for final classification. We conducted extensive experiments including various comparisons, and achieved promising results on the traffic scene dataset. To the best of our knowledge, we are the first to apply a spatial pyramid VLAD encoding scheme to the traffic scene recognition task.

The rest of this paper is organized as follows: Section 2 outlines some related research on visual representation and traffic scene recognition; Section 3 provides a detailed description of the proposed methods; implementation details and experimental results are provided in Section 4, followed by conclusion in Section 5.

## 2. Related Work

### 2.1. Visual Representation

In last decade, image recognition has advanced quickly because of extensive research [3][5][6]. The most popular conventional methods are the Bag of Visual Word representation [6] which is advantageous in representing local features into a single word to summarize the visual content. High dimensional encoding methods, such as Fisher Vector [8] and VLAD [7], were later proposed to reserve high-order information for better recognition.

In the last two years, deep convolutional neural networks have shown their effectiveness in visual representation learning, through various progresses in high-level vision tasks, e.g., image classification and scene parsing. These powerful CNN architectures have turned out to be effective for capturing intrinsic features, visual concepts or even semantic information. The advantage of CNN-based representation learning can be further leveraged by exploiting efficient coding scheme like VLAD. Some work has been published along this line, which usually consist of two steps, where a CNNs is first utilized to extract features from local patches and then the features are encoded and aggregated by conventional methods. For instance, Gong *et al.* [9] employed Vector of Locally Aggregated Descriptors (VLAD) for pooling multi-scale orderless global FC-features (MOP-CNN) for scene classification. Dixit *et al.* [5] designed a semantic Fisher Vector to aggregate features from multiple layers (both convolutional and fully-connected layers) of CNNs for scene recognition.

### 2.2. Traffic Scene Recognition

Automatic recognition of visual scene is an important issue in computer vision, which has a significant role in autonomous vehicles, traffic surveillance and management. Despite the daunting challenges in the recognition of traffic scenes, much research has been done, usually aimed at the automatic analysis of the road environment, or the detection and classification of possible objects in the traffic scene like vehicles and pedestrians. For example, Tang and Breckon [10] proposed a road classification scheme by exploiting color, texture and edge features from image sub-regions and applying a neural network for their classification. The approach was further developed by Mioulet et al. [11]. An urban scene understanding method was introduced by Ess et al. [12], in which the segmentation regions were labeled by a pre-training classifier.

Recently, a data mining methodology was published for driving-condition monitoring via CAN-bus data, based on the general data mining process [13]. In [14]，an auto-regressive stochastic processes was applied for the classification and retrieval of traffic video. Chen et al [15] put forward a novel concept of the atomic scene and established a framework for monocular traffic scene recognition, by decomposing a traffic scene into atomic scenes.

Compared with previous work, we leverage the representation learning capability of CNN by first extracting CNN features then applying spatial pyramid VLAD coding. Since it was introduced, VLAD [7] has been extensively applied as an efficient coding method to compactly represent images, particularly for a large scale dataset. By accumulating the residuals on each visual word concatenated into a single vector, VLAD achieves a reasonable balance between memory usage and performance [7]. The obvious downside of VLAD coding is its inability to preserve spatial information, which has not been stressed sufficiently. The most influential approach for encoding spatial information is the spatial pyramid matching scheme proposed by Lazebnik et al [16], which is a straightforward expansion of the bag-of-features representation, in which the histograms of the local features are gathered in each sub-region. A combination of a spatial pyramid and VLAD was introduced by Zhou et al [17]. We applied the similar methods of [4][17] by extracting deep activation features from local patches at multiple scales, and coding them with VLAD. While the emphasis of [4][17] was on scene classification and object classification, our focus is on the explicit abstraction of the traffic scene and the corresponding spatial information, which was no obviously evident in [4][17].

## 3. Methods

In this section, we will describe the main method we propose, which includes the extraction of region-based features based on the region proposal algorithm EdgeBoxes, VLAD encoding and Spatial Pyramid VLAD encoding. Fig.1 shows this system workflow.

### 3.1. Feature Extraction

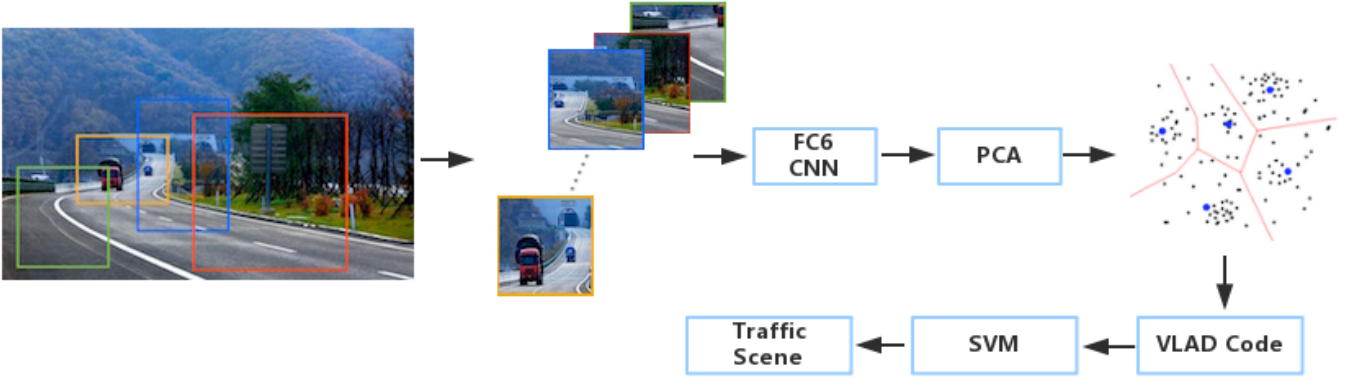

**FIGURE 1.** Illustration of the proposed method: Each window is generated by a region proposal algorithm and represented by FC6 features, the dimension reduction method: Principle Component Analysis (PCA) is applied, followed by K-means clustering for centroid learning (The blue dots). Traffic scene can then be classified with VLAD code and a SVM classifier.

In many vision tasks involving object detection as a component, region proposals have been regarded as standard practice. In our work, we will also start with a set of region proposals from the images. Among the approaches published, Edgeboxes [18] was applied in our work to produce high-quality region proposal, because of its high-level capability and computational efficiency. The pre-trained ImageNet model VGG16 was directly applied for feature extraction. For the task of traffic scene recognition, we used the Softmax Loss layer from the MatConvnet platform [19]. Based on our experience, we found that 1000 regions are adequate for image representation. We use the CNN model to extract CNN features from the first fully connection layer (FC6) for the 1000 high-quality region proposals by EdgeBoxes for each image. Since the algorithm of Edgeboxes provides ranking list for region proposal that have confidence values, the top 1000 region proposal have higher probabilities, which means most probably they contain a traffic scene. The number of clusters multiplied by the dimension of CNN features after PCA dimensionality reduction is the final dimension of the VLAD.

### 3.2. VLAD Encoding

VLAD encoding was proposed to encode a set of local feature vectors into a single compact vector. This encoding strategy achieves a reasonable balance between retrieval accuracy and memory footprint. The basic principles are as follows: Let $X = \{x_i\}_{i=1}^{n}$ be a set of local descriptors. Then a codebook $C = \{C_1,...,C_k\}$ of $k$ visual words can be learnt by the k-means algorithm. Each local descriptor $x_i$ can be assigned to its nearest visual word. For each visual word, the sum of the differences between the center and each local descriptor assigned to this center can be subsequently obtained. This can be expressed as:

$$\delta_j(X) = \sum_{i=1}^{N} a_j^i (c_j - x_i) \quad (1)$$

where $a_j^i$ is a binary assignment weight indicating if the local descriptors belongs to this visual words. Then the VLAD code can be obtained by concatenating the sum item of each visual word:

$$v(X) = \left[\delta_1^T(X), \delta_2^T(X),...,\delta_m^T(X)\right] \quad (2)$$

Following the steps of VLAD, the codewords learning with k-means clustering will be performed with the number of clusters set to 16. Based on the codewords learnt by k-means, we perform the VLAD coding on the CNN features obtained after PCA, the final dimension of VLAD is the number of clusters multiplied by the dimension of the CNN features after PCA dimensionality.

### 3.3. Spatial Pyramid VLAD

Although VLAD encoding performs well in preserving local features, VLAD coding does not preserve spatial information. To deal with this problem, several recent papers have proposed spatial pyramid VLAD as a solution [4][16]. In this research, we implemented this method for traffic scene classification. As shown in Fig.2, the level of the spatial pyramid is 2x2. Regions are allocated into each spatial grid, with an assignment determined by the distribution of the regions centers.

As has being pointed out in [7], appropriate dimension reduction on original features would further improve the performance of VLAD encoding. Therefore, we use Principal Component Analysis (PCA) [20] to perform dimensionality reduction on the CNN features extracted from these regions. However, as the number of features is large, training conventional PCA on all of the features would be unrealistic. We firstly randomly select 220K sampled regions for training and reduced the CNN features from 4096 dimensions to 256. Then we perform PCA on all of the remaining features.

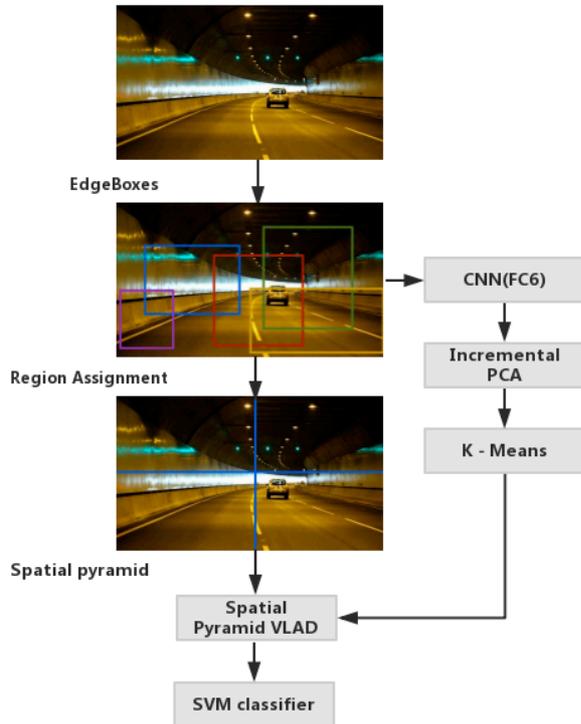

**FIGURE 2.** VLAD encoding with a spatial pyramid: The image is divide using a 2x2 level spatial pyramid.

## 4. Experiments

In the following section, we will first introduce the traffic scene dataset, and then the experimental set up will be briefly outlined including performance comparison, followed by the details of experiments on the database for traffic scene recognition.

### 4.1. Dataset preparation

Our dataset contains 2000 images assigned to 10 categories of traffic scenes with 200 images belonging to each category (Fig.3): bridges, gas station, highway, indoor parking, outdoor parking, roundabout, toll station, traffic jams, train station and tunnel. The average size of each image is approximately 450*500 pixels. The images of the 10 categories were obtained by us from both the Google image search engine as well as personal photographs. Each category of scenes was split randomly into two separate sets of images, 135 for training and the rest for testing.

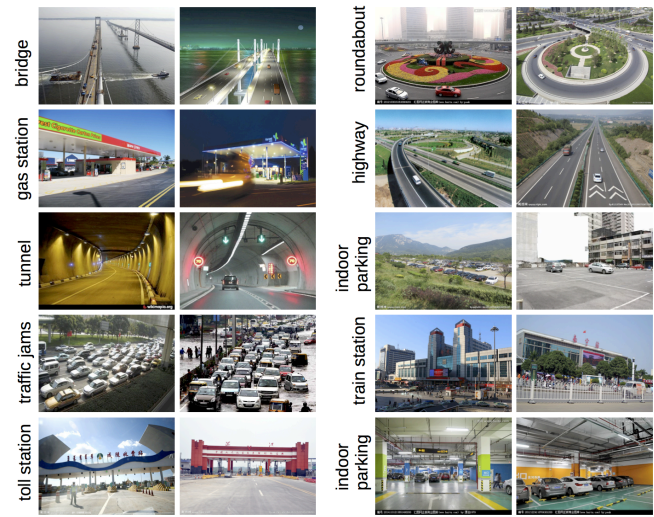

**FIGURE 3.** Image samples from the traffic scene dataset

### 4.2. Deep Learning Model

All of the models have been implemented on the MatConvNet deep learning framework [19]. We employed VGG 16 with the network pre-trained on ImageNet and extract the CNN features directly and applied a classifier SVM for the final classification.

### 4.3. VLAD Encoding

Our experiments were conducted under the Linux operating system. The incremental PCA was implemented in Matlab. The obtained dimensionality reduced features, k-means Clustering and VLAD encoding were realized in Matlab using the VLFeat toolbox [21]. As traffic scene prediction can be considered a classification problem, a SVM linear classifier was utilized from the LIBSVM toolbox [22].

### 4.4. Traffic Scene Recognition

We evaluated our method on the traffic scene dataset, which is split into training and testing sets of 1350 and 650 instances respectively to evaluate the system performance. The images within each class have large variations in backgrounds and images angles.

We followed the Spatial Pyramid VLAD encoding of CNN features as previously explained, and applied a SVM classifier for the final prediction. Specifically, the VGG16 model was utilized for feature extraction. The region proposal algorithm EdgeBoxes was applied on each image, and FC6 features were then extracted for each region. The VLAD encoding was accomplished after PCA dimensionality reduction and codewords learning with clustering. More details about the experiment procedure and three comparative settings are described as follows:

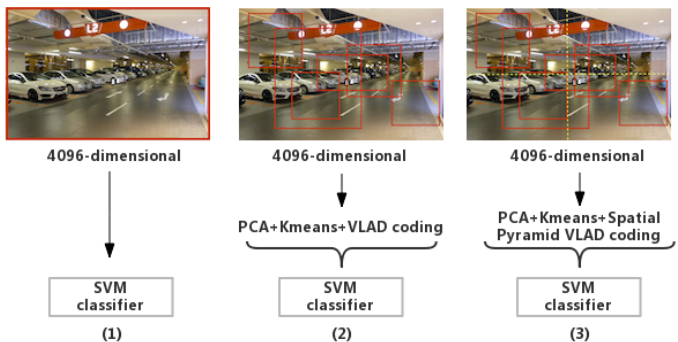

**FIGURE 4.** VLAD encoding with a spatial pyramid: The image is divided with a 2x2 level spatial pyramid. Overview of three comparative experiments settings: (1) level 1, 4096-dimensional CNN feature for the entire image, directly applied for traffic scene classification. (2) level 2, extract CNN feature from 1000 region proposals and VLAD coding them with a codebook of 16 centers. (3) based on level 2, we added spatial pyramid information before VLAD encoding.

(1) CNN features.

To evaluate the stand-alone performance of VGG16, the CNN features from the first fully connected layers (FC6) corresponding to each image are directly applied for traffic scene classification as a comparative baseline. The accuracy is 93.54%, which implies that CNN features for traffic scene recognition are sufficient. The images were directly input to the CNN without candidate objects extraction by a region proposal algorithm.

(2) VLAD coding.

In these tests, 1000 boxes for each image were generated by the region proposal algorithm Edgeboxes, which was represented by CNN features of 256 dimensionality and then VLAD coding was applied to the 16 learnt codewords. The accuracy increases up to 95.85%, and the spatial pyramid has not been taken into account.

(3) Deep Spatial pyramid VLAD coding.

Finally, to test the influence on the overall performance of the spatial pyramid VLAD encoding, we added spatial pyramid encoding, and concatenated the VLAD codes of each pyramid into one representation. Experimental results showed that adding the spatial pyramid does improve the overall performance and the accuracy is 96.15%. The confusion matrices in Fig.4 further proves the significance of the spatial pyramid VLAD encoding.

Confusion matrix of level 1

Confusion matrix of level 2

Confusion matrix of level 3

**FIGURE 4.** The confusion matrices of traffic scene database with three comparative settings

## 5. Conclusion

In this paper, we presented a novel traffic scene recognition system with demonstrated satisfactory performance on a traffic scene dataset of 10 categories. Experimental results indicate that information from local patches and the global contextual information are significant contributing factors to improve the performance of traffic scene recognition. This is substantiated by our reimplementation of the Vector of Locally Aggregated Descriptors (VLAD) on top of a spatial pyramid for CNN features to catch local information and global spatial information simultaneously. Experiments were conducted for different settings, with results confirmed that the VLAD codes brings performance gains for traffic scene recognition. The beneficial effect of spatial pyramids has also been demonstrated with performance enhancement.